\documentclass[10pt,twocolumn,letterpaper]{article}
\usepackage[T1]{fontenc}
\usepackage{newunicodechar}
\newunicodechar{–}{--}\newunicodechar{‐}{-}\newunicodechar{’}{'}
\usepackage[pagenumbers]{cvpr}

\usepackage{cuted} 

\usepackage{currfile} 

\usepackage{caption} 

\usepackage{graphicx}
\usepackage{booktabs}
\usepackage{amsmath}
\usepackage[varbb]{newtxmath}
\usepackage{array,multirow}
\usepackage{xcolor}
\usepackage{array,multirow,longtable}
\usepackage{placeins}
\definecolor{cvprblue}{rgb}{0.21,0.49,0.74}
\usepackage[pagebackref,breaklinks,colorlinks,allcolors=cvprblue]{hyperref}
\usepackage[capitalize,noabbrev]{cleveref}
\graphicspath{{figures/}}
\newcommand{\pp}{\,\mathrm{pp}}
\newcommand{\Full}{\textsc{Full40000}}
\newcommand{\Ufive}{\textsc{U5000}}
\newcommand{\Rfive}{\textsc{Random5000}}
\newcommand{\Loss}{\textsc{Loss5000}}
\newcommand{\Ulow}{\textsc{U1250}}
\newcommand{\Rlow}{\textsc{Random1250}}
\newcommand{\Query}{\textsc{D4-Query}}
\newcommand{\Uniform}{\textsc{D4-Uniform}}

\def\confName{CVPR}
\def\confYear{2027}
\title{Where Do Human-to-Robot Transfer Gains Come From?}
\author{Jiachen Li\textsuperscript{1}\quad Shihao Li\textsuperscript{1}\quad Soovadeep Bakshi\textsuperscript{2}\quad Longteng Fan\textsuperscript{3}\\
Dongmei Chen\textsuperscript{1}\quad Long Chen\textsuperscript{3}\quad Zhuo Xu\textsuperscript{3}\\[4pt]
\textsuperscript{1}University of Texas at Austin\quad \textsuperscript{2}Independent\quad \textsuperscript{3}General Intelligent Machines, Inc}

\newcommand{\sref}[1]{\cref{#1}}\newcommand{\mref}[1]{\cref{#1}}
\begin{document}
\maketitle
\begin{abstract}
Human data can support robot policy pretraining, but gains attributed to a data recipe may also reflect subset size, training variability, or downstream adaptation. In an HPT implementation on a 20-task MetaWorld suite with three training blocks, a constrained random 5,000-recording subset matches most of the task-informed subset's gain over the 40,000-recording pool at 20,000 updates (3.69 of 4.29 percentage points), leaving equal-size selection differences of 0.60 points at 5,000 recordings and 0.45 points at 1,250 recordings, with wide paired intervals. Across 12 comparisons, task orderings repeat across seeds for several contrasts, but a fixed predictor built from the other two seeds trails the mean or difficulty reference in at least one held-out fold for every comparison. Neither mechanism probe shows its predicted positive interaction: task-slot mismatch yields an interaction of $-0.0347$ points after adaptation, and in a six-task observation experiment the observation-by-pretraining interaction changes sign between seeds ($-18.29$ and $+1.04$ points) while full-state pretraining gains are positive in both ($+6.37$ and $+2.31$ points). This retrospective study shows how equal-size and capacity controls, seed-level reporting, and targeted interventions change the interpretation of apparent transfer gains.
\end{abstract}

\section{Introduction}
\begin{figure*}[t]
\centering
\includegraphics[width=\linewidth]{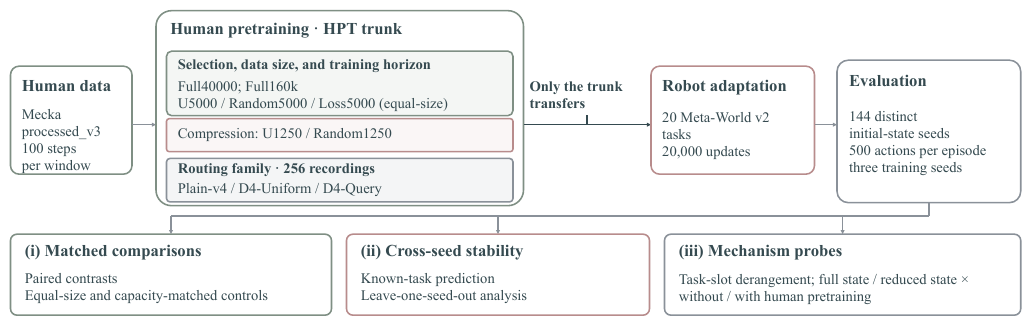}
\caption{\textbf{One transfer pipeline, three questions.} Human recordings pretrain the HPT trunk under data recipes from three separate protocol families; only the trunk transfers to robot adaptation on 20 Meta-World tasks, and every model is evaluated on 144 initial states per task across three training seeds. The analyses then ask whether a recipe beats a matched control, whether its task-level effects repeat across seeds, and whether targeted interventions support the proposed mechanism.}
\label{fig:workflow}
\end{figure*}
Heterogeneous pretraining offers a way to use human experience for robot learning, but the value of a particular data recipe is an attribution problem. A selected subset can outperform a larger pool because of which recordings it contains, how often its windows are revisited, or how a fixed optimization budget interacts with the resulting training distribution. A positive average also need not identify which tasks benefit reproducibly. These distinctions matter when deciding whether a data-selection criterion or transfer mechanism has actually been validated.

We study these questions using an implementation of Heterogeneous Pre-trained Transformers (HPT; \citealt{wang2024scaling}) and a fixed 20-task MetaWorld suite \citep{yu2020meta}. The empirical program first compared human-data recipes and later examined smaller subsets, task-conditioned routing, and two proposed mechanisms. This paper consolidates that program retrospectively, evaluating each experimental family against its own controls and documenting earlier exploratory variants in the appendix.

The central example is the difference between a subset comparison and a selection comparison. At 20,000 human pretraining updates across three training seeds, the equal-size constrained random subset matches most of the task-informed 5,000-recording subset's gain over the 40,000-recording pool (3.69 of 4.29 percentage points), leaving an equal-size selection difference of 0.60 points (95\% interval $[-13.15,14.35]$). Under fourfold compression to 1,250 recordings, the equal-size selection difference is 0.45 points ($[-5.87,6.78]$), and the task-informed subset trails 5,000 random recordings by 3.34 points ($[-16.02,9.33]$), whose $-16.02$-point lower bound falls short of the $-2$-point noninferiority margin.

Task-level analysis separates rank repeatability from gain calibration. Predicting each known task's effect in one seed from the other two seeds yields negative rank correlations in all three folds for 5,000-recording selection ($-0.211$, $-0.478$, $-0.427$) and query routing ($-0.228$, $-0.185$, $-0.102$). Several other contrasts, including both subset-versus-full contrasts and 1,250-recording selection, keep positive rank correlations in all three folds, while the fixed two-seed predictor has higher MSE than the mean or difficulty reference in at least one fold of each of the 12 comparisons.

Finally, two interventions test candidate mechanisms. After 20,000 adaptation updates, mismatching task slots leaves the query-routed model's success unchanged, where a selective loss was predicted; the interaction with the uniform control is $-0.0347$ points. In a six-task observation experiment, pretraining gains with full state are positive in both seeds ($+6.37$ and $+2.31$ points), and masking environment-state coordinates, predicted to enlarge them, gives interactions of $-18.29$ and $+1.04$ points.

Our contribution is an empirical separation of three questions: \textbf{(i)} whether a data recipe has an average advantage over an appropriate control; \textbf{(ii)} whether its task-specific effects persist across training repetitions; and \textbf{(iii)} whether a particular intervention supports the proposed explanation. We answer each in this setting (\cref{fig:workflow}) with matched comparisons and an analysis rebuilt from saved success counts, and we give the protocol detail needed to apply the same separation to other transfer claims.

\section{Related work}
\label{sec:related}

\paragraph{Heterogeneous pretraining and human-to-robot transfer.}
Heterogeneous Pre-trained Transformers (HPT) use embodiment-specific input interfaces and a shared policy trunk to learn from heterogeneous proprioceptive and visual data \citep{wang2024scaling}.
Our experiments use one HPT implementation and tasks from Meta-World \citep{yu2020meta}.
Human data have been used in several different ways: MimicPlay learns latent plans from human play to guide robot control \citep{wang2023mimicplay}, while EgoMimic combines aligned human and robot demonstrations in policy learning \citep{kareer2025egomimic}.
More recent work studies how hand-label quality, motion differences, and embodiment-specific processing affect cotraining on everyday human videos \citep{li2026matters}.
Visual representations pretrained on human video, such as R3M \citep{nair2022r3m}, VIP \citep{ma2022vip} and masked visual pretraining \citep{radosavovic2023real}, draw on egocentric corpora such as Ego4D \citep{grauman2022ego4d}, while cross-embodiment datasets and generalist policies pool data from many robots \citep{o2024open,team2024octo}.
These settings establish useful context for our observation and initialization controls, but differ from our frozen source pool, simulator tasks, and training schedules.

\paragraph{Data curation and mixtures.}
Data curation has objectives beyond reducing the number of demonstrations.
DemInf estimates the contribution of trajectories to state--action mutual information, balancing action diversity and predictability \citep{hejna2025robot}.
Re-Mix learns domain weights through distributionally robust optimization for robot pretraining \citep{hejna2024re}.
DataMIL uses datamodels and a task-specific surrogate to select data for downstream policy success \citep{dass2026datamil}.
Gradient-based attribution scores training examples through gradient similarity, retraining-fitted datamodels or random-projection estimators \citep{pruthi2020estimating,ilyas2022datamodels,park2023trak}, and LESS selects instruction-tuning data by gradient alignment with a target set \citep{xia2024less}, the closest analogue of our utility U. \citet{belkhale2023data} characterize demonstration quality through action divergence and transition diversity, and task gradients can conflict in multi-task training \citep{yu2020gradient}.
These approaches distinguish data utility from simple notions of source quality.
A recent controlled study of VLA scaling also reports negative transfer from naive data mixtures and limits of several regularization strategies \citep{wang2026rethinking}.
Complementing these selector designs, we examine how matched subset sizes, training repetitions, and interventions change the interpretation of one project's data recipes.

\paragraph{Empirical evaluation and uncertainty.}
An extensive study of offline robot imitation found sensitivity to dataset quality, algorithm choices, and stopping criteria \citep{mandlekar2021matters}.
In reinforcement learning, \citet{henderson2018deep} documented substantial variability and reproducibility difficulties, and \citet{agarwal2021deep} showed how conclusions drawn from aggregate point estimates can differ from uncertainty-aware analysis.
\citet{bouthillier2021accounting} show that data sampling, initialization and hyperparameter choices add variance that benchmark comparisons must account for.
These concerns also matter when supervised policy training is evaluated through stochastic rollouts.
We distinguish variation across training seeds from evaluation uncertainty conditional on a trained model, and examine whether task-level effects retain their ordering or magnitude across seeds.
Together with matched controls, the leave-one-seed-out analysis on the fixed task suite separates average gains from task-level repeatability.

\section{Experimental design}
\label{sec:methods}

We analyze a sequence of completed experiments on one human-to-robot transfer pipeline. The experiments share an HPT-style transformer and robot adaptation recipe, but address different questions with different human-data pools. We therefore compare arms within each protocol family. The present synthesis is retrospective; individual controls were specified before their corresponding runs.

\subsection{Data, model, and evaluation}

The human data are the \texttt{processed\_v3} Mecka \texttt{flagship} and \texttt{freeform} recordings, exported to the study's \texttt{neutral\_v2} representation. Project records associate this historical source with the EgoVerse ecosystem~\citep{punamiya2026egoverse}; the fixed study export is distinct from the complete current public release. Each target contains 128 values: ego-compensated keypoints from both hands and two aperture scalars. A window samples 100 steps at stride three from a 300-frame segment. Windows start at raw-frame offsets $0,300,600,\ldots$; incomplete tails are discarded. Four RGB observations, at window steps $0,33,66,99$, accompany each target sequence. We sample uniformly over the selected recordings' windows, so recording selection and training exposure are distinct units.

The human policy uses an ImageNet-initialized ResNet-18 visual stem~\citep{he2016deep}, a 16-block, width-256, eight-head HPT transformer~\citep{wang2024scaling}, and a task-specific action head. The shared trunk has 12,644,352 parameters. Human training minimizes mean smooth-$L_1$ loss with $\beta=0.05$ over the 100-by-128 target. The visual stem, trunk, and human head are optimized jointly. Each chain starts with a fresh trunk initialization. Only the trunk transfers to the robot policy; the human visual stem and head do not transfer.

Robot adaptation uses a fixed, custom suite of 20 Meta-World v2 tasks~\citep{yu2020meta}, listed in \sref{app:protocol}. Each task has 100 accepted scripted-expert demonstrations, split into 90 training and 10 validation episodes. Robot observations contain RGB features from a frozen ImageNet ResNet-18 and a 39-dimensional state vector; the observation intervention below changes only the latter. The policy predicts eight four-dimensional actions. We fine-tune the full robot policy for 20,000 updates, with task-specific input stems and action heads. \sref{tab:hyperparameters} lists the optimization settings, and \sref{app:protocol} gives normalization and temporal conventions.
For each model and task, evaluation uses 144 distinct initial-state seeds and 500 actions per episode. Each episode constructs a fresh environment and renderer after seeding the random-number generators. An episode succeeds if the environment reports success at any step; evaluation continues to the full horizon. Arms share the same ordered initial-state seeds within a comparison. The four parallel execution shards contain 36 different episodes each. Task success is the mean of 144 binary outcomes, and the suite score weights tasks equally. Training seeds are the replication unit for between-training comparisons, and we report all three seed-level values.

Writing $m_s=T^{-1}\sum_{t=1}^{T}d_{t,s}$, the reported three-block mean and descriptive interval are
\begin{equation}
\begin{gathered}
 \bar m=\tfrac13\sum_{s=1}^{3}m_s,\\
 \bar m\ \pm\ t_{0.975,2}\,\frac{\operatorname{SD}(m_1,m_2,m_3)}{\sqrt{3}},\qquad t_{0.975,2}=4.30265273.
\end{gathered}
\label{eq:tinterval}
\end{equation}
Each interval summarizes the three training blocks for a single contrast, unadjusted across the 12 contrasts. The two-point margins recorded in the compression and mechanism studies are exploratory decision thresholds.

\subsection{Selection, data size, and training horizon}

The large-pool study fixes 40,000 complete human recordings containing 311,595 windows. Full40000 uses this pool for 20,000 human updates. U5000, Random5000, and Loss5000 each select exactly 5,000 recordings subject to a total-window interval of $[38{,}560,39{,}338]$. All three training seeds reuse the same selected set for each method. To compare training horizons against Full40000, Full160k trains on the full pool from a fresh initialization with a 160,000-update cosine schedule.

\paragraph{Gradient-alignment utility U.}
Let $g_w$ be the trunk gradient of the unweighted human loss for window $w$, evaluated at a frozen reference model $\Theta$. A fixed subsampled randomized Hadamard transform $P$ maps it to 16,384 dimensions. We form
\begin{equation}
 h_w=\frac{P g_w}{\lVert P g_w\rVert_2},\qquad
 \bar h_i=\frac{1}{n_i}\sum_{w\in i}h_w,\qquad
 a_{it}=\tfrac{1}{2}\,\bar h_i^{\top}\bigl(q_{t,A}+q_{t,B}\bigr),
 \label{eq:u-score}
\end{equation}
where $n_i$ is the recording's window count. The recording mean is not renormalized. Each unit query $q_{t,b}$ is a normalized mean of 20 unit episode-gradient vectors. The two banks use 40 additional scripted-expert demonstrations per task, disjoint from the first 100 adaptation demonstrations and from evaluation rollouts. They use task-specific query heads fitted with the reference trunk frozen. U is therefore task-supervised gradient alignment, not prediction uncertainty.

Each column $a_{\cdot t}$ is converted to a midrank percentile over the complete 40,000-recording pool. The scalar utility $u_i$ is the equal-task mean of these percentiles. Binary selection solves
\begin{equation}
 \begin{aligned}
 \max_{x\in\{0,1\}^N}\ \ & \textstyle\sum_i x_i u_i\\
 \text{s.t.}\ \ & \textstyle\sum_i x_i=K,\qquad W_{\min}\leq\sum_i x_i n_i\leq W_{\max}.
 \end{aligned}
 \label{eq:selection}
\end{equation}
The baseline named Random uses random permutation priorities as utilities in the same constrained optimization. It is a constrained random-priority selector, not a uniform draw over all feasible subsets. Loss minimizes the mean frozen-reference loss per recording under the same constraints; it uses no robot query gradients. Scoring uses the original export-header normalization, while subsequent training uses one common human normalizer across arms.

The compression study keeps this recipe but sets $K=1{,}250$ and the window interval to $[9{,}640,9{,}834]$. U1250 uses one frozen subset. Random1250 uses three priority draws, paired one-to-one with training seeds 1--3. Thus its three blocks jointly vary the draw and training initialization. Random5000 outcomes are reused from the completed large-pool study.

\subsection{Task-slot routing and interventions}

\paragraph{Routing family.}
Plain-v4, D4-Uniform, and D4-Query share a previously selected set of 256 recordings and 1,901 windows from a 4,096-recording candidate pool; its archived v4 selector combines U with three timestep-weighted gradient scores (\sref{app:protocol}). D4 adds 20 rank-four residual slots, low-rank adapters in the style of LoRA \citep{hu2021lora}, to the last transformer block's MLP output projection. During human training, each batch of 64 windows assigns three or four windows to each slot. D4-Query maximizes the batch's recorded task affinities under these capacities; D4-Uniform permutes the same capacity labels independently of the model and sampler random streams. At transfer, one slot is merged into the ordinary trunk for each robot task; robot adaptation contains no slot router. The three-seed comparison includes the development seed and two subsequent seeds.

\paragraph{Slot correspondence.}
For seed 3, two additional robot-training arms use a fixed derangement of the task-to-slot mapping, one for D4-Query and one for D4-Uniform. Correctly mapped controls are reused. With $S$ denoting suite success, the diagnostic is
\begin{equation}
 I_{\mathrm{slot}}=(S_{Q,c}-S_{Q,m})-(S_{U,c}-S_{U,m}),
 \label{eq:slot}
\end{equation}
where $c$ and $m$ denote correct and mismatched slots. It measures the effect of correspondence after 20,000 adaptation updates.

\paragraph{Observation availability.}
A separate four-cell experiment crosses a trained human trunk versus its saved random initialization with full state versus RGB plus eight proprioceptive values. Cells A/B use full state without/with Plain-v4 human pretraining; C/D use reduced state without/with pretraining. Six tasks and training seeds 2 and 3 were fixed before the pilot. All cells receive exactly the same task-and-seed-specific robot nontrunk initialization. B's 12 trained robot models are reused, but all four cells are evaluated anew on a common fresh reset grid.

Reduced state is implemented by zeroing object and goal coordinates after normalization, retaining the original 39-dimensional stem. RGB, current and previous hand position and gripper values, task identity, and rendered goal markers remain available. The primary endpoint is 20,000 updates; 5,000 updates is a readout of the same schedule. For each seed, the interaction is
\begin{equation}
 I_{\mathrm{obs},s}=\frac{1}{6}\sum_t
 \big[(S_{D,s,t}-S_{C,s,t})-(S_{B,s,t}-S_{A,s,t})\big].
 \label{eq:obs}
\end{equation}
Section~\ref{sec:limitations} discusses the scope of both probes.

\section{Data recipes under controlled budgets}
\label{sec:selection}
\subsection{Subset competitiveness and equal-size selection}
Table~\ref{tab:arms} reports every training block for the selection, compression, and routing families. Comparisons are made within a family. In the 40,000-recording experiment, \Ufive{} reaches 54.26\% mean success and \Rfive{} reaches 53.66\%, compared with 49.97\% for \Full{}. The differences against the full pool are therefore $+4.29$ and $+3.69\pp$, respectively. The \Ufive{} minus \Rfive{} differences are $+1.28$, $-5.24$, and $+5.76\pp$; their mean is $+0.60\pp$.

\begin{figure}[t]
\centering
\includegraphics[width=\linewidth]{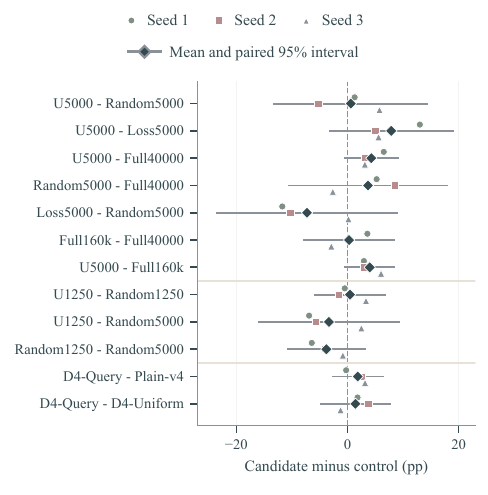}
\caption{Within-family paired effects in percentage points. Small markers show the three training-block differences; the aggregate marker and interval show their mean and descriptive 95\% paired $t$ interval ($n=3$). Every interval crosses zero.}
\label{fig:paired}
\end{figure}

The equal-size random subset thus matches most of the selected subset's gain over \Full{}, and the paired 95\% interval for selection itself is $[-13.15,14.35]$. Figure~\ref{fig:paired} shows the block values behind each interval; they measure training-seed variability for the one fixed random subset.

\begin{table}[t]
\caption{Task-macro success rates (\%) by training block. Standard deviation is across the three block values. The three experiment families use separate protocols. Random5000 is reused in the compression comparisons and is listed once.}
\label{tab:arms}
\centering\small
\begin{tabular}{lrrrrr}
\toprule
Condition & Seed 1 & Seed 2 & Seed 3 & Mean & SD \\
\midrule
\multicolumn{6}{l}{\textit{Selection / training horizon}} \\
Full40000 & 53.51 & 46.91 & 49.48 & 49.97 & 3.33 \\
U5000 & 60.03 & 50.14 & 52.60 & 54.26 & 5.15 \\
Random5000 & 58.75 & 55.38 & 46.84 & 53.66 & 6.14 \\
Loss5000 & 47.01 & 45.10 & 47.01 & 46.38 & 1.10 \\
Full160k & 57.08 & 47.12 & 46.56 & 50.25 & 5.92 \\
\midrule
\multicolumn{6}{l}{\textit{Compression (new conditions)}} \\
U1250 & 51.84 & 49.76 & 49.34 & 50.31 & 1.34 \\
Random1250 & 52.33 & 51.25 & 46.01 & 49.86 & 3.38 \\
\midrule
\multicolumn{6}{l}{\textit{Task routing}} \\
Plain-v4 & 53.92 & 49.20 & 43.72 & 48.95 & 5.11 \\
D4-Uniform & 51.88 & 48.09 & 48.12 & 49.36 & 2.18 \\
D4-Query & 53.68 & 51.84 & 46.88 & 50.80 & 3.52 \\
\bottomrule
\end{tabular}

\end{table}

\subsection{A source-loss control and longer training}
The frozen source-loss selector (\Loss{}) reaches 46.38\% mean success, trailing \Rfive{} by $7.28\pp$ on average across blocks ($-11.74$, $-10.28$, and $+0.17\pp$; paired interval $[-23.42,8.86]$).

Extending full-pool human pretraining from 20,000 to 160,000 updates (from a fresh initialization with a 160,000-update cosine schedule) changes mean success from 49.97\% to 50.25\% ($+0.29\pp$, interval $[-7.78,8.36]$), remaining below both 5,000-recording subsets (53.66\% and 54.26\%).

\subsection{Compression to 1,250 recordings}
At 1,250 recordings, \Ulow{} reaches 50.31\% mean success and \Rlow{} reaches 49.86\%. The matched block differences ($-0.49$, $-1.49$, and $+3.33\pp$) average $+0.45\pp$ with interval $[-5.87,6.78]$, below the planned $+2\pp$ threshold for a useful gain.

Relative to the reused \Rfive{} endpoints, \Ulow{} changes mean success by $-3.34\pp$ (interval $[-16.02,9.33]$); the lower bound, $-16.02\pp$, falls short of the $-2\pp$ noninferiority margin. \Rlow{} trails \Rfive{} in all three blocks, with a $-3.80\pp$ mean difference.

\subsection{Task-conditioned routing as a separate comparison}
The smaller-pool routing family compares Plain-v4 (48.95\%), capacity-matched \Uniform{} (49.36\%), and \Query{} (50.80\%). \Query{} leads Plain-v4 by $+1.85\pp$ and the capacity-matched \Uniform{} by $+1.44\pp$; both paired intervals span zero. The \Query{} minus \Uniform{} difference is $+1.81\pp$ on development seed 1, $+3.75\pp$ on seed 2, and $-1.25\pp$ on seed 3.

\section{What repeats across training seeds?}
\label{sec:crossseed}
\subsection{A fixed known-task prediction test}
For a candidate $a$, control $b$, task $t$, and training block $s$, let $d_{t,s}=100(p_{a,t,s}-p_{b,t,s})$ be the success-rate difference in percentage points. For each held-out block $s$, the same-task predictor uses the other two blocks:
\begin{equation}
\widehat{d}^{\,\mathrm{task}}_{t,s}=\frac{1}{2}\sum_{r\neq s} d_{t,r}.
\end{equation}
We compare it with an overall-mean predictor fitted on those same training blocks, a fixed quadratic difficulty reference, and a zero-effect predictor. The difficulty reference regresses the training-block task effect on $1$, $q_t$, and $q_t(1-q_t)$, where $q_t$ is the control success fraction averaged only over the two training blocks. Held-out success rates are never used to construct a prediction.

For each of the 12 comparisons, the 20 tasks and three held-out blocks produce 60 predictions. We report mean squared error (MSE, $\mathrm{pp}^2$), mean absolute error, and foldwise Spearman correlation; positive MSE improvement (reference MSE minus same-task MSE) favors same-task prediction. Zero predictions count as incorrect directions when the observed effect is nonzero, and observed zero effects remain in MSE. The analysis plan was saved after the original results had been inspected and before these diagnostics were calculated, making this an exploratory reanalysis of known tasks; its 720 predictions share endpoints and overlapping training folds.

\begin{figure*}[t]
\centering
\includegraphics[width=\linewidth]{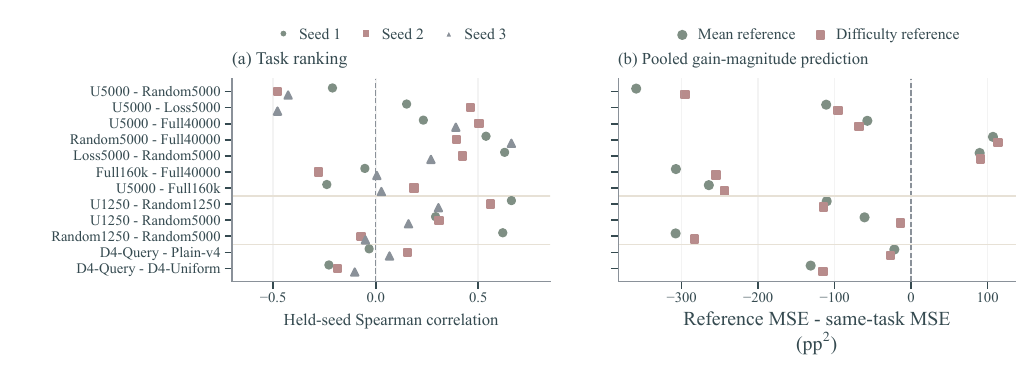}
\caption{Task-rank repeatability and gain-magnitude prediction across all 12 comparisons for the two-seed same-task predictor; positive MSE improvement favors same-task prediction. No comparison beats both references in all three folds. Protocol families are separated.}
\label{fig:crossseed}
\end{figure*}

\subsection{Selection and routing effects are negatively rank-correlated across seeds}
For \Ufive{} minus \Rfive{}, the three held-out rank correlations are $-0.211$, $-0.478$, and $-0.427$. Same-task prediction has higher MSE than both the mean and difficulty references in every fold, with pooled MSE improvements of $-359.38$ and $-295.75\,\mathrm{pp}^2$, and matches the observed sign on eight of 36 nonzero effects.

The Query--Uniform comparison shows the same pattern: foldwise rank correlations of $-0.228$, $-0.185$, and $-0.102$, pooled MSE improvements of $-131.30$ and $-115.37\,\mathrm{pp}^2$, and 10 correct signs among 28 nonzero effects.

\subsection{Positive rank correlations coexist with uncalibrated gain magnitudes}
Several contrasts retain positive task-ordering correlations across all three folds. \Rfive{} minus \Full{} reaches correlations of $+0.537$, $+0.394$, and $+0.661$, with pooled MSE improvements of $+107.03$ and $+113.20\,\mathrm{pp}^2$, and \Loss{} minus \Rfive{} (which shares the \Rfive{} endpoint) also has positive correlations in all folds and positive pooled MSE improvements. In held-out seed 2 for the first and seed 3 for the second, at least one simple reference has lower MSE.

The compression contrast further separates ordering from magnitude. \Ulow{} minus \Rlow{} yields correlations $+0.661$, $+0.558$, and $+0.306$, with 21 correct directions among 31 nonzero effects. Yet its pooled MSE is worse than both references, by $110.32$ and $114.54\,\mathrm{pp}^2$, respectively. Similarly, all three macro differences for \Ufive{} minus \Full{} are positive, as are its three rank correlations, while pooled gain-magnitude prediction is worse than the simple references.

Across all 12 comparisons, the fixed two-seed predictor trails at least one of the two simple references in at least one held-out fold (\cref{tab:crossall}; task heatmaps in \sref{app:results}).

\begin{table*}[t]
\caption{All cross-seed comparisons. Positive MSE improvement favors the fixed same-task predictor over the specified reference. The final column counts held-out folds in which its MSE is lower than both the mean and difficulty references.}
\label{tab:crossall}
\centering\scriptsize
\begin{tabular}{lrrrrrr}
\toprule
Comparison & $\rho_1$ & $\rho_2$ & $\rho_3$ & $\Delta$MSE: mean & $\Delta$MSE: difficulty & Folds \\
\midrule
U5000 - Random5000 & -0.211 & -0.478 & -0.427 & -359.4 & -295.8 & 0/3 \\
U5000 - Loss5000 & +0.151 & +0.462 & -0.478 & -111.0 & -95.8 & 1/3 \\
U5000 - Full40000 & +0.232 & +0.503 & +0.390 & -57.2 & -67.9 & 1/3 \\
Random5000 - Full40000 & +0.537 & +0.394 & +0.661 & +107.0 & +113.2 & 2/3 \\
Loss5000 - Random5000 & +0.628 & +0.422 & +0.269 & +89.6 & +90.8 & 2/3 \\
Full160k - Full40000 & -0.053 & -0.279 & +0.006 & -307.5 & -254.8 & 0/3 \\
U5000 - Full160k & -0.238 & +0.187 & +0.028 & -264.5 & -244.0 & 0/3 \\
U1250 - Random1250 & +0.661 & +0.558 & +0.306 & -110.3 & -114.5 & 1/3 \\
U1250 - Random5000 & +0.292 & +0.309 & +0.160 & -60.9 & -13.7 & 1/3 \\
Random1250 - Random5000 & +0.619 & -0.071 & -0.050 & -307.9 & -283.2 & 1/3 \\
D4-Query - Plain-v4 & -0.032 & +0.154 & +0.067 & -21.9 & -27.1 & 1/3 \\
D4-Query - D4-Uniform & -0.228 & -0.185 & -0.102 & -131.3 & -115.4 & 0/3 \\
\bottomrule
\end{tabular}

\end{table*}

\begin{figure*}[t]
\centering
\includegraphics[width=\linewidth]{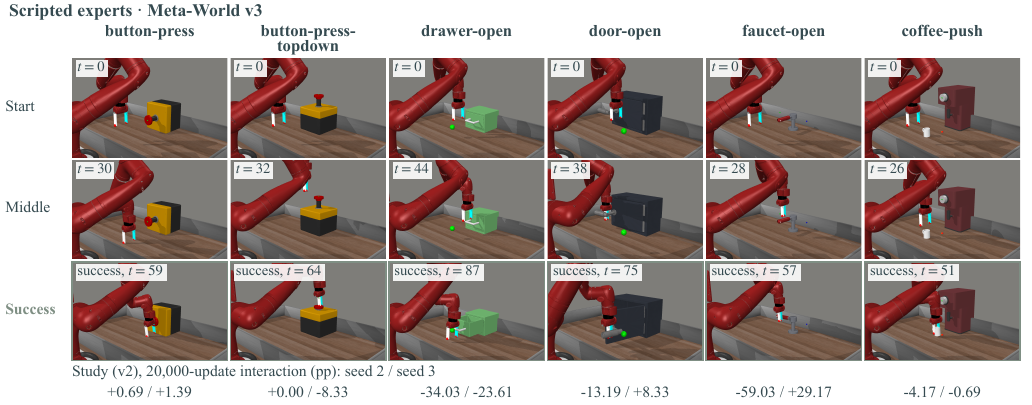}
\caption{\textbf{The six observation-probe tasks.} Illustrative keyframes of scripted expert demonstrations, rendered with one fixed camera in the current Meta-World release (v3 tasks); the study used the v2 tasks, and these frames do not show its trained policies. Under each task, the study's 20,000-update observation-by-pretraining interaction for seeds 2 and 3, in percentage points (\cref{tab:obstasks}).}
\label{fig:keyframes}
\end{figure*}
\section{Two bounded mechanism probes}
\label{sec:mechanisms}
\begin{figure}[t]
\centering
\includegraphics[width=\linewidth]{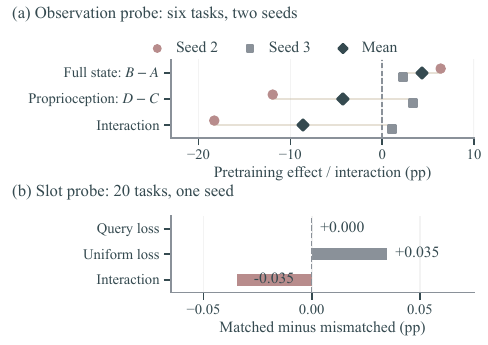}
\caption{Separate mechanism probes. The slot intervention uses one parent/downstream seed and one fixed derangement after 20,000 adaptation updates. The observation intervention uses six tasks and two training seeds; each seed's full-input gain, reduced-input gain, and interaction are shown.}
\label{fig:mechanisms}
\end{figure}
\subsection{Does query routing depend on correct task-slot correspondence?}
The first probe uses seed-3 Query and Uniform human parents and the same downstream seed. For each parent, it compares the correct task-slot assignment with one frozen derangement, followed by the same 20,000-update downstream adaptation. The correct-assignment endpoints are reused; the two mismatched conditions are newly trained. The principal contrast is the loss from mismatch in Query minus the loss from mismatch in Uniform.

Query achieves 46.8750\% with either assignment, and Uniform changes from 48.1250\% to 48.0903\% (one episode across 2,880 episodes). The interaction is therefore $-0.0347\pp$, where the plan predicted positive selective damage with recorded two-point targets for the Query mismatch loss and for the difference of mismatch losses; task-level shifts in both directions offset across the suite.

The probe measures correspondence that still matters after 20,000 adaptation updates; weak slot information, downstream overwriting, and dominance of the shared trunk all fit the observation.

\subsection{Does reduced numerical state increase the value of human pretraining?}
The second probe tests an observation-by-pretraining interaction on six fixed tasks and training seeds 2 and 3. All four cells retain RGB encoded by the same frozen ImageNet ResNet-18 and robot proprioception. The full-input cells receive 39 numerical state coordinates; reduced-input cells preserve eight proprioceptive coordinates and mask the remaining coordinates after normalization. Cells A/C omit human policy pretraining but keep the ImageNet-pretrained visual encoder; B/D use the matched human parent. B models are reused, and all four cells are evaluated on the same new reset states.

At the primary 20,000-update endpoint, the mean full-input pretraining gain $B-A$ is $+4.34\pp$ ($+6.37$ and $+2.31\pp$ across seeds), whereas the reduced-input effect $D-C$ averages $-4.28\pp$ ($-11.92$ and $+3.36\pp$). The difference-in-differences
\begin{equation}
 I_{\mathrm{obs}}=(D-C)-(B-A)
\end{equation}
averages $-8.62\pp$, with seed values $-18.29$ and $+1.04\pp$: the interaction that masking was predicted to make consistently positive changes sign between seeds (task-level values in \cref{tab:obstasks}).

\begin{table}[t]
\caption{Observation probe at 20,000 updates: all six tasks and both training seeds. Values are percentage-point effects, with interaction $(D-C)-(B-A)$.}
\label{tab:obstasks}
\centering\footnotesize
\begin{tabular}{lrrrr}
\toprule
Task & Seed & B - A & D - C & Interaction \\
\midrule
button-press-topdown & 2 & +0.00 & +0.00 & +0.00 \\
button-press-topdown & 3 & +0.00 & -8.33 & -8.33 \\
button-press & 2 & +0.00 & +0.69 & +0.69 \\
button-press & 3 & +0.00 & +1.39 & +1.39 \\
coffee-push & 2 & +0.00 & -4.17 & -4.17 \\
coffee-push & 3 & +0.00 & -0.69 & -0.69 \\
door-open & 2 & -2.78 & -15.97 & -13.19 \\
door-open & 3 & -8.33 & +0.00 & +8.33 \\
drawer-open & 2 & +0.00 & -34.03 & -34.03 \\
drawer-open & 3 & +25.69 & +2.08 & -23.61 \\
faucet-open & 2 & +40.97 & -18.06 & -59.03 \\
faucet-open & 3 & -3.47 & +25.69 & +29.17 \\
\bottomrule
\end{tabular}

\end{table}

The reduced-input cells reach 68.98\% and 64.70\% mean success, so the masked condition remains learnable in aggregate. \Cref{fig:keyframes} shows the six tasks with their per-seed interactions. Local floors and ceilings still matter: the drawer task is difficult in the reduced-input condition, coffee pushing is at zero in the full-input condition, and several button settings saturate. At the diagnostic 5,000-update checkpoint, interaction values are $-15.28$ and $+14.12\pp$, again with opposite signs. Removing one task at a time leaves the 20,000-update average interaction negative (from $-10.56$ to $-4.58\pp$), and for the slot probe it keeps the Query-minus-Uniform interaction between $-0.1462$ and $+0.2924\pp$. The complete cells, task definitions, and conditional evaluation interval are reported in the appendix.

\section{Discussion and limitations}
\label{sec:limitations}
\paragraph{What the controls reveal.}
The equal-size random control changes the selection story: Random5000 matches $+3.69\pp$ of the $+4.29\pp$ gain of U5000 over Full40000, leaving $+0.60\pp$ for selection, and U1250 leads Random1250 by only $+0.45\pp$. The capacity-matched control does the same for routing: Query leads Plain-v4 by $+1.85\pp$ but Uniform by $+1.44\pp$, so a comparison with Plain-v4 alone would credit routing structure to query-informed allocation. The transferable lesson is to align each control with the proposed attribution: equal recording and window budgets for selection, matched capacity for routing, and matched visual preprocessing and initialization for observation interventions.

\paragraph{Different evidence answers different questions.}
Mean gains, repeatable task ordering, and calibrated magnitudes come apart in these data: U5000 exceeds Full40000 in all three blocks ($+4.29\pp$ mean) with positive task-rank correlations, and U1250 minus Random1250 has rank correlations of $+0.306$ to $+0.661$, yet the fixed two-seed predictor's pooled MSE is worse than the simple references in both contrasts. The mechanism probes add a third kind of evidence: their null and sign-changing interactions count against the specific predictions they test and leave other mechanisms open.

\paragraph{Scope and statistical precision.}
This retrospective synthesis evaluates one HPT implementation on a fixed 20-task Meta-World suite across protocol families that share reused endpoints. Primary comparisons use three training blocks with unadjusted paired intervals that are too wide to establish superiority or equivalence. Random5000 uses one fixed subset draw, Random1250 couples three subset draws to the three training seeds, D4 seed 1 served as a development seed, Full160k changes the update horizon and cosine schedule together, and the cross-seed analysis evaluates one fixed two-seed predictor on known tasks across overlapping folds. Saturated tasks reduce the information in some contrasts. Development screens of five other auxiliary objectives (D1--D3, D5, D6) scored below their designated controls, and only task routing (D4) advanced to the three-seed and mismatch studies (\sref{app:history}).

\paragraph{Mechanism and observation boundaries.}
The slot probe tests one derangement on seed 3 after 20,000 adaptation updates, measuring the combined effect of initial slot content and downstream fine-tuning. The observation probe tests a specific coordinate mask on six tasks across two seeds while retaining RGB and task identity, with task-level floors and ceilings and a reset-bootstrap interval ($[-10.19,-7.06]\pp$) that quantifies evaluation variation conditional on the fixed trained models rather than retraining variation across seeds.

\paragraph{Reproducibility and artifact scope.}
The accompanying package rebuilds the paper's statistics, tables, and figures, including the observation bootstrap, from saved success counts and episode outcomes without running a simulator or model. End-to-end retraining additionally requires the human recordings, normalization arrays, and training environment, which are outside the package; some historical environments and large archives were removed during the project. The release candidate holds numerical simulator outcomes and study descriptions and excludes human videos, trajectories, recording identifiers, normalization arrays, and model weights (\sref{app:reproduction}).

\section{Conclusion}
In these HPT/MetaWorld experiments, an equal-size random subset matches most of the task-informed subset's gain over the 40,000-recording pool ($+3.69$ of $+4.29\pp$), leaving $+0.60\pp$ for selection at 5,000 recordings and $+0.45\pp$ at 1,250, and the capacity-matched control leaves query routing $+1.44\pp$ ahead. Task orderings repeat across seeds for several contrasts and are negatively correlated for selection at 5,000 recordings and for query routing, while the fixed two-seed predictor trails a simple reference in at least one fold of every comparison. Neither mechanism probe shows its predicted positive interaction. Attribution, repeatability, and mechanism each need their own comparison, and keeping them apart changes what apparent human-to-robot transfer gains can be taken to mean.

\section*{LLM Usage Statement}
AI tools, including Claude Code, OpenAI Codex, and Google Gemini, were used to assist with coding, manuscript editing, and formatting. All implementations, results, and conclusions were independently reviewed and verified by the authors. AI tools were not used for data generation, theoretical development, or proofs. The authors take full responsibility for the content.

{
    \small
    \bibliographystyle{ieeenat_fullname}
    \bibliography{references}
}

\clearpage
\appendix
\renewcommand{\thesection}{S\arabic{section}}\renewcommand{\thefigure}{S\arabic{figure}}\renewcommand{\thetable}{S\arabic{table}}\renewcommand{\theequation}{S\arabic{equation}}
\setcounter{section}{0}\setcounter{figure}{0}\setcounter{table}{0}\setcounter{equation}{0}
\renewcommand{\theHequation}{S\arabic{equation}}
\begin{center}{\Large\bf Appendix}\end{center}
\section{Protocol and implementation details}
\label{app:protocol}

\subsection{Populations and task suite}

The human export was finalized on 13 August 2026. Its source header identifies Mecka \texttt{processed\_v3}, native Zarr recordings from \texttt{flagship} and \texttt{freeform}, and the \texttt{neutral\_v2} output schema. The eligible export contains 57,478 recordings and 447,373 windows. Eligibility requires the recorded head, wrist, end-effector, hand-keypoint, and front-video channels and at least 300 raw frames. Arrays are truncated to the declared true frame count. Six duplicate recording identifiers across sources use the first occurrence, with flagship preceding freeform. The large-pool comparison fixes a 40,000-recording subset before scoring; two unreadable first-frame recordings were replaced through the original seed-1 ordering before the final pool was frozen. The small-pool routing experiments instead use the original C4096 candidate roster (31,132 windows).

The 20 tasks are assembly, basketball, bin-picking, box-close, button-press-topdown, button-press, button-press-wall, coffee-button, coffee-pull, coffee-push, dial-turn, disassemble, door-close, door-lock, door-open, door-unlock, hand-insert, drawer-close, drawer-open, and faucet-open, all with the \texttt{-v2} suffix. This custom suite is the project's 21-task manifest without button-press-topdown-wall. The observation pilot uses button-press, button-press-topdown, drawer-open, door-open, faucet-open, and coffee-push. That subset was selected through a task-geometry and visual-observability audit before its outcomes were collected.

\subsection{Training architecture and numerical settings}

\begin{table*}[t]
\centering
\footnotesize\setlength{\tabcolsep}{3pt}
\caption{Training settings used by the ordinary human and robot policies. D4 changes only the specified human trunk layer.}
\label{tab:hyperparameters}
\begin{tabular}{@{}p{0.27\linewidth}p{0.33\linewidth}p{0.32\linewidth}@{}}
\toprule
Setting & Human pretraining & Robot adaptation \\
\midrule
Trunk & \multicolumn{2}{p{0.67\linewidth}}{16 blocks; width 256; 8 heads; drop path 0.1} \\
Batch size & 64 windows & 256 windows \\
Optimizer & AdamW & AdamW \\
Peak learning rate & $10^{-4}$ & $10^{-5}$ \\
Adam $\beta_1,\beta_2$ & $(0.9,0.95)$ & $(0.9,0.999)$ \\
Adam $\epsilon$ & $10^{-8}$ & $10^{-7}$ \\
Weight decay & $10^{-2}$ & $10^{-4}$ \\
Warmup & none & 1,000 updates, linear from zero \\
Cosine endpoint & zero & $10^{-6}$ \\
Update horizon & 20,000; Full160k: 160,000 & 20,000 \\
Smooth-$L_1$ $\beta$ & 0.05 & 0.05 \\
Target shape & $100\times128$ & $8\times4$ \\
\bottomrule
\end{tabular}
\end{table*}

The human visual stem takes four $224\times224$ RGB frames, applies ImageNet mean/std normalization, uses the ResNet-18 layer-4 feature map, projects it to width 256, and cross-attends with 16 learned latent tokens (eight heads of dimension 64). Its backbone is trainable during human pretraining. The HPT-style human action head has hidden widths 256 and 128, dropout enabled, and no final tanh. The optimizer receives all policy parameters. The human loader shuffles the flat window roster each epoch and drops an incomplete final batch. The trunk is freshly initialized for each recorded chain seed, while the frozen scoring reference is a separate previously trained checkpoint.

The fixed human training normalizer is fitted per source on the original candidate-manifest training windows: 120,500 flagship and 2,992,700 freeform timesteps, totaling $31{,}132\times100$. It is reused across the selection, full-pool, and compression arms. Standard deviations have a $10^{-8}$ floor. Gradient and loss scoring instead retain the original export-header constants, fitted on the eligible export universe.

Robot images use the \texttt{view\_1} wrist-mounted camera at $128\times128$ resolution. A frozen ImageNet ResNet-18 supplies $49\times512$ cached features. The robot policy uses image and state cross-attention stems with 16 latents each, one block, eight heads of dimension 64, and modality dropout 0.1. Input projections have widths $512\rightarrow128\rightarrow256$ for image features and $39\rightarrow128\rightarrow256$ for state. The action head has widths $256\rightarrow256\rightarrow128\rightarrow32$, dropout enabled, and final tanh. Modality embeddings are enabled, domain embeddings disabled, and trunk tokens are mean pooled. The robot trunk, stems, and head are all fine-tuned at the same learning rate.

Cached expert transitions follow the source implementation's one-step action shift: state/image at one stored index is paired with the next stored action. Training uses observation horizon one and action horizon eight; sequences do not cross episode boundaries or pad incomplete action tails. Inference uses up to four observations and executes eight-action chunks. Raw actions are clipped to $[-1,1]$. The episode-level validation split uses seed 233 and ratio 0.1. The limits-based state/action normalizer is fitted to all 100 cached fit demonstrations, including the ten validation demonstrations; it is then common to the compared arms. Final evaluation initial states are separate from these demonstrations.

\subsection{Selection and query construction}

Query heads were fitted for 20,000 robot updates at seed 1 while keeping $\Theta$'s trunk frozen. For each task, accepted expert indices $[100,120)$ and $[120,140)$ form banks A and B; the first 100 accepted episodes supply robot fitting. The query loss uses evaluation-mode policy behavior: history up to four, action chunks of eight, clipped actions, and mean smooth-$L_1$ loss over valid action coordinates. Each episode's projected trunk gradient is normalized before averaging; each bank mean is normalized again. The two banks contain 800 demonstrations across the 20 tasks. All subsequent U, v4, and D4 uses reuse these query vectors.

Projection pads the 12,644,352-dimensional trunk gradient to $2^{24}$ and uses a fixed SRHT output dimension $2^{14}$. Human window directions are individually normalized, recording means are not, and bank A/B scores receive equal weight. Midrank percentiles are computed over the entire candidate pool. The mixed-integer solver imposes binary inclusion, exact cardinality, and two window inequalities, with zero relative optimality gap.

The actual 5,000-recording totals are 38,560 windows for U, 39,241 for Random, and 39,338 for Loss. Loss is the unweighted mean HPT-Huber loss over each window's 100 timesteps and 128 coordinates, averaged equally over windows within each recording. Random5000 uses priority seed 1. At $K=1{,}250$, U retains 9,640 windows; the three Random draws (125001, 125002, 125003) retain 9,674, 9,716, and 9,831 windows. These draws are paired with training seeds 1, 2, and 3, respectively.

The v4 score used only in the small-pool family is
\begin{equation}
 f_{it}=\frac{1}{2}R_t(U_{it})+\frac{1}{6}\bigl[R_t(T_{it})+R_t(B_{it})+R_t(E_{it})\bigr],
\end{equation}
where $R_t$ ranks over all C4096 records. T, B, and E are the same gradient-alignment construction with distinct nonnegative timestep weights inside the human loss. T compares per-dimension energy of target differences in a frozen transferable-motion subspace and its orthogonal residual. B weights deviation from an edge-padded four-step moving average. E weights affine-camera-compensated RAFT-small flow residuals relative to hand motion. B and E use clipped 5th--95th-percentile normalization; E uses a $10^{-6}$ denominator offset. These timestep weights apply only to the archived scoring features. The selector ranks $f_{\cdot t}$ again per task, averages tasks, and uses \mref{eq:selection} with $K=256$ and window bounds $[1{,}901,1{,}939]$. Plain-v4 and both D4 arms all use its same complete 1,901-window selection.

\subsection{Exact interventions}

For D4, the modified layer is block 15's $1024\rightarrow256$ MLP output projection, using zero-based indexing. For task $t$, its weight is $W+B_tA_t$, where $A_t\in\mathbb{R}^{4\times1024}$ and $B_t\in\mathbb{R}^{256\times4}$. All 20 down matrices begin with the same Gaussian draw (standard deviation 0.02, dedicated seed 1), all up matrices begin at zero, and the shared base remains trainable. The 102,400 added parameters disappear as separate parameters when a slot is merged for downstream training.

At zero-based optimizer step $s$, every slot has capacity three, with one additional window assigned to slots $(4s+j)\bmod20$, $j=0,1,2,3$. Query routing solves the resulting $64\times64$ maximum-affinity assignment using the selected recordings' full-C4096 percentile v4 scores. Uniform routing uses a dedicated NumPy random generator seeded at 1 to permute the identical capacity labels. No new gradient scoring occurs during D4 routing. The mismatch experiment applies one fixed 20-task derangement to the exported slot identity; task data, robot nontrunk initialization, training seed, update count, and evaluation states remain matched. Its complete mapping is listed in \texttt{protocol/slot-mismatch-mapping.csv} in the accompanying analysis package.

In the observation pilot, reduced state retains zero-based coordinates $\{0,1,2,3,18,19,20,21\}$: current and previous hand position and gripper values. Coordinates $4{:}18$ and $22{:}39$ are set to zero after the original per-dimension normalization and before the unchanged state stem, in both training and inference. All cells copy the same saved B-arm robot nontrunk tensors and normalizer. A/C use the corresponding human parent's saved step-zero trunk; B/D use its 20,000-update trunk. Both 5,000- and 20,000-update readouts belong to the same 20,000-update schedule. A fresh set of 144 reset seeds per task is shared across cells, training seeds, and readouts: 864 unique task--reset pairs and 13,824 evaluated episodes in total.

\paragraph{Conditional reset uncertainty.}
The observation report uses 5,000 paired bootstrap draws with NumPy generator seed 2026110702. Within each of the six fixed tasks, each draw resamples 144 reset indices with replacement. The same index multiplicities are applied to all four cells, both training seeds, and both checkpoint readouts. Effects are averaged equally across tasks and then across the two fixed training seeds; the 2.5th and 97.5th percentiles give the reported interval. This resampling estimates reset uncertainty conditional on the fitted models and chosen tasks.

\subsection{Runtime continuity and artifact availability}

The retained runtime record for the November experiments reports Python 3.10.20, PyTorch 2.1.1 with CUDA 12.1, torchvision 0.16.1, NumPy 1.24.3, SciPy 1.10.1, \texttt{mujoco-py} 2.1.2.14, and MuJoCo 2.1.0. The study uses a pinned HPT/Meta-World source; the upstream HPT transformer reference is commit \texttt{549433f7}, and the human stem and training assembly are study-specific.

The October model archives and the original working runtime were deleted before the November follow-up; lightweight results and source/configuration copies survived. The follow-up restored a compatible runtime from a surviving environment and reconstructed the original first-100 expert streams, matching all 20 saved raw-content and nonimage-array references and all four preserved image-array references; for the other tasks, image-array reconstruction relies on raw-content identity and the frozen encoder implementation. Previously completed outcome records were reused and enter each comparison once, and newer model archives are recorded in the follow-up delivery receipts.

\clearpage
\section{Complete effect and reproducibility summaries}
\label{app:results}
\subsection{Statistical definitions}
For a trained model, task success is the fraction of 144 independently reset episodes in which the environment's official success flag is reached at least once during a horizon of 500 actions. Methods share the task-specific reset-seed schedule for paired evaluation. The task macro for a training block averages task success fractions with equal weight, and the training seeds are the repeated units for the three-block paired intervals.

\Cref{eq:tinterval} of the main text defines the three-block mean and its descriptive interval.

Table~\ref{tab:allpaired} reports all paired contrasts across the full task suite, and the accompanying CSV files retain all three block values.
\begin{table*}[t]
\caption{All 12 within-family paired contrasts. Differences and interval endpoints are in percentage points.}
\label{tab:allpaired}
\centering\small
\begin{tabular}{lrrrrr}
\toprule
Comparison & Seed 1 & Seed 2 & Seed 3 & Mean & 95\% interval \\
\midrule
U5000 - Random5000 & +1.28 & -5.24 & +5.76 & +0.60 & [-13.15, +14.35] \\
U5000 - Loss5000 & +13.02 & +5.03 & +5.59 & +7.88 & [-3.19, +18.96] \\
U5000 - Full40000 & +6.53 & +3.23 & +3.12 & +4.29 & [-0.51, +9.10] \\
Random5000 - Full40000 & +5.24 & +8.47 & -2.64 & +3.69 & [-10.51, +17.89] \\
Loss5000 - Random5000 & -11.74 & -10.28 & +0.17 & -7.28 & [-23.42, +8.86] \\
Full160k - Full40000 & +3.58 & +0.21 & -2.92 & +0.29 & [-7.78, +8.36] \\
U5000 - Full160k & +2.95 & +3.02 & +6.04 & +4.00 & [-0.38, +8.39] \\
U1250 - Random1250 & -0.49 & -1.49 & +3.33 & +0.45 & [-5.87, +6.78] \\
U1250 - Random5000 & -6.91 & -5.62 & +2.50 & -3.34 & [-16.02, +9.33] \\
Random1250 - Random5000 & -6.42 & -4.13 & -0.83 & -3.80 & [-10.78, +3.18] \\
D4-Query - Plain-v4 & -0.24 & +2.64 & +3.16 & +1.85 & [-2.70, +6.40] \\
D4-Query - D4-Uniform & +1.81 & +3.75 & -1.25 & +1.44 & [-4.83, +7.70] \\
\bottomrule
\end{tabular}

\end{table*}

\subsection{Cross-seed prediction details}
The saved endpoint table has 660 rows. Sixty rows repeat the Random5000 endpoints reused by the compression study. Deduplicating identical arm--seed--task keys leaves 600 unique endpoints. The reproduction script verifies identical integer success counts for repeated keys before constructing task matrices. Each endpoint has 144 evaluation episodes. Correlations use average ranks for ties, calculated from integer count differences so that tied or zero effects remain exact.

The difficulty reference uses the mean control fraction in the two training seeds, with features $(1,q_t,q_t(1-q_t))$. Its least-squares response is the candidate-minus-control effect averaged over the same two seeds, fitted across the 20 tasks. Both the response and features exclude the held-out seed. The fitted value at those training-defined features predicts the same known task in the held-out seed. The predictor and references were fixed before the held-out metrics were computed.

Pooled MSE and MAE average all 60 task--fold errors, foldwise rank correlations are reported per held-out seed, and direction accuracy uses every nonzero observed effect as its denominator, counting zero predictions as misses. All actual-zero cases remain included in MSE and MAE.
\Cref{tab:crossall} of the main text lists all 12 comparisons.

\subsection{Observation and slot cells}
Tables~\ref{tab:obscells} and \ref{tab:slotcells} retain the experimental cells behind the interactions. The observation study records two checkpoints along the same training trajectories: a diagnostic readout at 5,000 updates and the primary endpoint at 20,000. The reported conditional 95\% interval for the mean 20,000-update interaction is $[-10.19,-7.06]\pp$, which reflects reset-state evaluation noise for the fixed trained models. Stability across retraining is read from the two training-seed interactions, $-18.29$ and $+1.04\pp$, which change sign.

Under leave-one-task-out omission, the slot probe's Query mismatch loss ranges from $-0.2193$ to $+0.3289\pp$ and the Query-minus-Uniform interaction from $-0.1462$ to $+0.2924\pp$; the observation probe's mean 20,000-update interaction ranges from $-10.56$ to $-4.58\pp$.

\begin{table*}[t]
\caption{Observation-by-pretraining cells at the diagnostic and primary readouts. All cells retain the same frozen ImageNet visual encoder and RGB input; the pretraining factor is human policy pretraining.}
\label{tab:obscells}
\centering\small
\begin{tabular}{rlllrrr}
\toprule
Step & Arm & State & Human & Seed 2 & Seed 3 & Mean \\
\midrule
5000 & A & Full39 & No & 71.99 & 77.31 & 74.65 \\
5000 & B & Full39 & Yes & 70.37 & 77.08 & 73.73 \\
5000 & C & Proprio8 & No & 67.59 & 50.23 & 58.91 \\
5000 & D & Proprio8 & Yes & 50.69 & 64.12 & 57.41 \\
20000 & A & Full39 & No & 75.00 & 75.81 & 75.41 \\
20000 & B & Full39 & Yes & 81.37 & 78.12 & 79.75 \\
20000 & C & Proprio8 & No & 75.35 & 62.62 & 68.98 \\
20000 & D & Proprio8 & Yes & 63.43 & 65.97 & 64.70 \\
\bottomrule
\end{tabular}

\end{table*}

\begin{table*}[t]
\caption{Observation pretraining effects and interactions, in percentage points. The 5,000-update readout is diagnostic; 20,000 updates is the primary endpoint.}
\label{tab:obseffects}
\centering\small
\begin{tabular}{rlrrr}
\toprule
Step & Comparison & Seed 2 & Seed 3 & Mean \\
\midrule
5000 & B - A & -1.62 & -0.23 & -0.93 \\
5000 & D - C & -16.90 & +13.89 & -1.50 \\
5000 & Interaction & -15.28 & +14.12 & -0.58 \\
20000 & B - A & +6.37 & +2.31 & +4.34 \\
20000 & D - C & -11.92 & +3.36 & -4.28 \\
20000 & Interaction & -18.29 & +1.04 & -8.62 \\
\bottomrule
\end{tabular}

\end{table*}

\begin{table*}[t]
\caption{Task-slot probe after 20,000 downstream updates. Correct-assignment endpoints are reused. This probe contains one parent/downstream training seed and one fixed derangement.}
\label{tab:slotcells}
\centering\small
\begin{tabular}{lrrr}
\toprule
Arm & Seed & Successes / episodes & Success (\%) \\
\midrule
D4-Query & 3 & 1350 / 2880 & 46.875 \\
D4-Query-Mismatched & 3 & 1350 / 2880 & 46.875 \\
D4-Uniform & 3 & 1386 / 2880 & 48.125 \\
D4-Uniform-Mismatched & 3 & 1385 / 2880 & 48.090 \\
\bottomrule
\end{tabular}

\end{table*}
\FloatBarrier

\subsection{Task-level mechanism contrasts}
\Cref{tab:slottasks} and \mref{tab:obstasks} list the task-level mechanism effects: observation interactions vary across the six tasks and two seeds, while the $-0.0347\pp$ slot interaction averages offsetting positive and negative task effects.

\begin{table*}[t]
\caption{Slot probe at 20,000 updates: all 20 tasks, training seed 3. Loss is correct minus mismatched success; interaction is Query loss minus Uniform loss. Values are percentage points.}
\label{tab:slottasks}
\centering\small
\begin{tabular}{lrrr}
\toprule
Task & Query loss & Uniform loss & Interaction \\
\midrule
assembly & +0.000 & +0.000 & +0.000 \\
basketball & +0.000 & +0.000 & +0.000 \\
bin-picking & +0.000 & +0.000 & +0.000 \\
box-close & +1.389 & -0.694 & +2.083 \\
button-press-topdown & +0.000 & +0.000 & +0.000 \\
button-press & +0.000 & +0.000 & +0.000 \\
button-press-wall & +4.167 & +4.167 & +0.000 \\
coffee-button & +0.000 & +0.000 & +0.000 \\
coffee-pull & +0.000 & +0.694 & -0.694 \\
coffee-push & +0.000 & +0.000 & +0.000 \\
dial-turn & +2.083 & +0.000 & +2.083 \\
disassemble & +0.000 & +0.000 & +0.000 \\
door-close & +0.000 & +0.000 & +0.000 \\
door-lock & -0.694 & +0.000 & -0.694 \\
door-open & -6.250 & +0.000 & -6.250 \\
door-unlock & -1.389 & -2.778 & +1.389 \\
drawer-close & +0.694 & -0.694 & +1.389 \\
drawer-open & +0.000 & +0.000 & +0.000 \\
faucet-open & +0.000 & +0.000 & +0.000 \\
hand-insert & +0.000 & +0.000 & +0.000 \\
\bottomrule
\end{tabular}

\end{table*}
\FloatBarrier

\clearpage
\subsection{Every task and training block}
The heatmaps show all 20 tasks for all 12 core contrasts, including ties, sign changes, and task-level heterogeneity across the suite. The corresponding machine-readable endpoint and effect tables are included in the analysis package.
\begin{figure*}[p]
\centering
\includegraphics[width=.97\linewidth,height=.78\textheight,keepaspectratio]{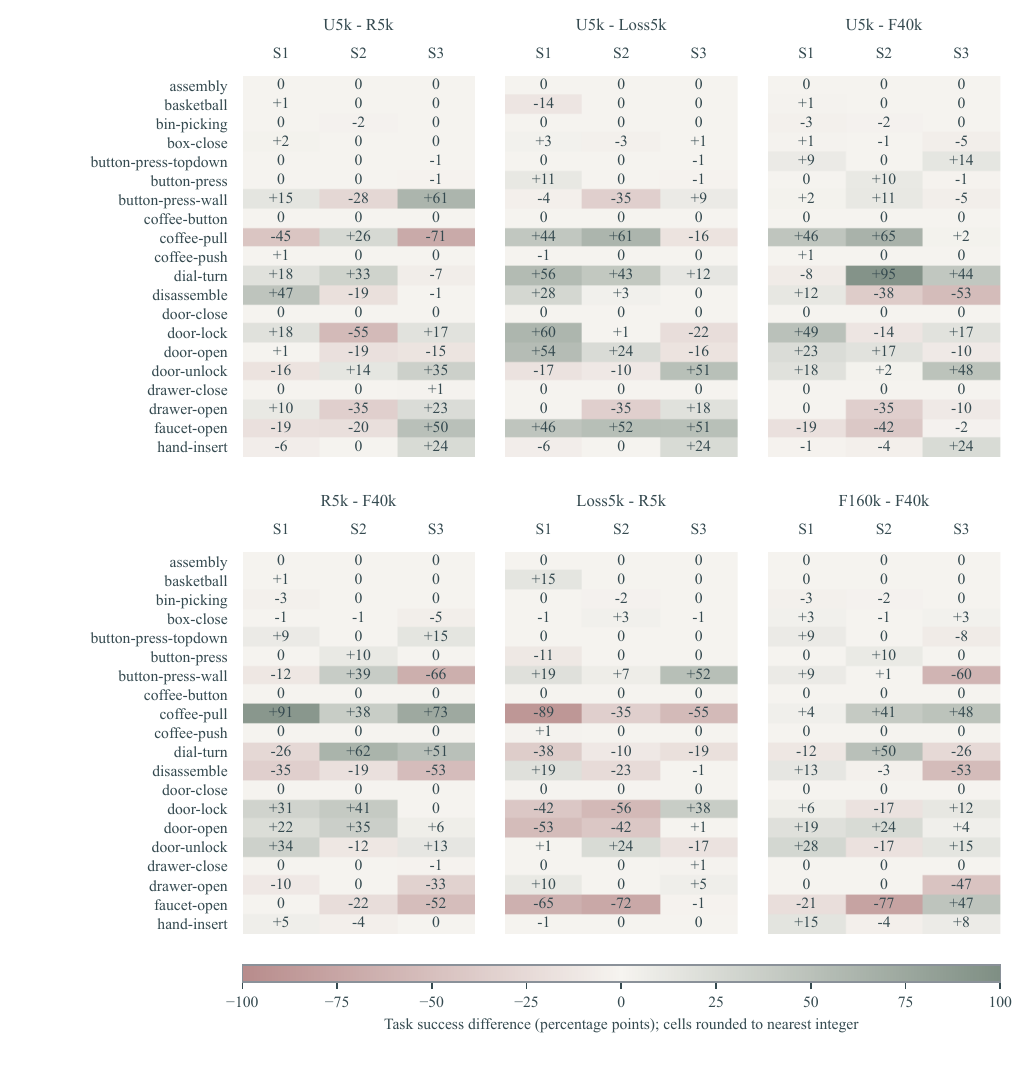}
\caption{Task-by-seed effects for the first six comparisons in the fixed analysis order. Color and numeric annotations show percentage-point differences. Zero denotes an exact tied success count.}
\label{fig:heat1}
\end{figure*}
\clearpage
\begin{figure*}[p]
\centering
\includegraphics[width=.97\linewidth,height=.83\textheight,keepaspectratio]{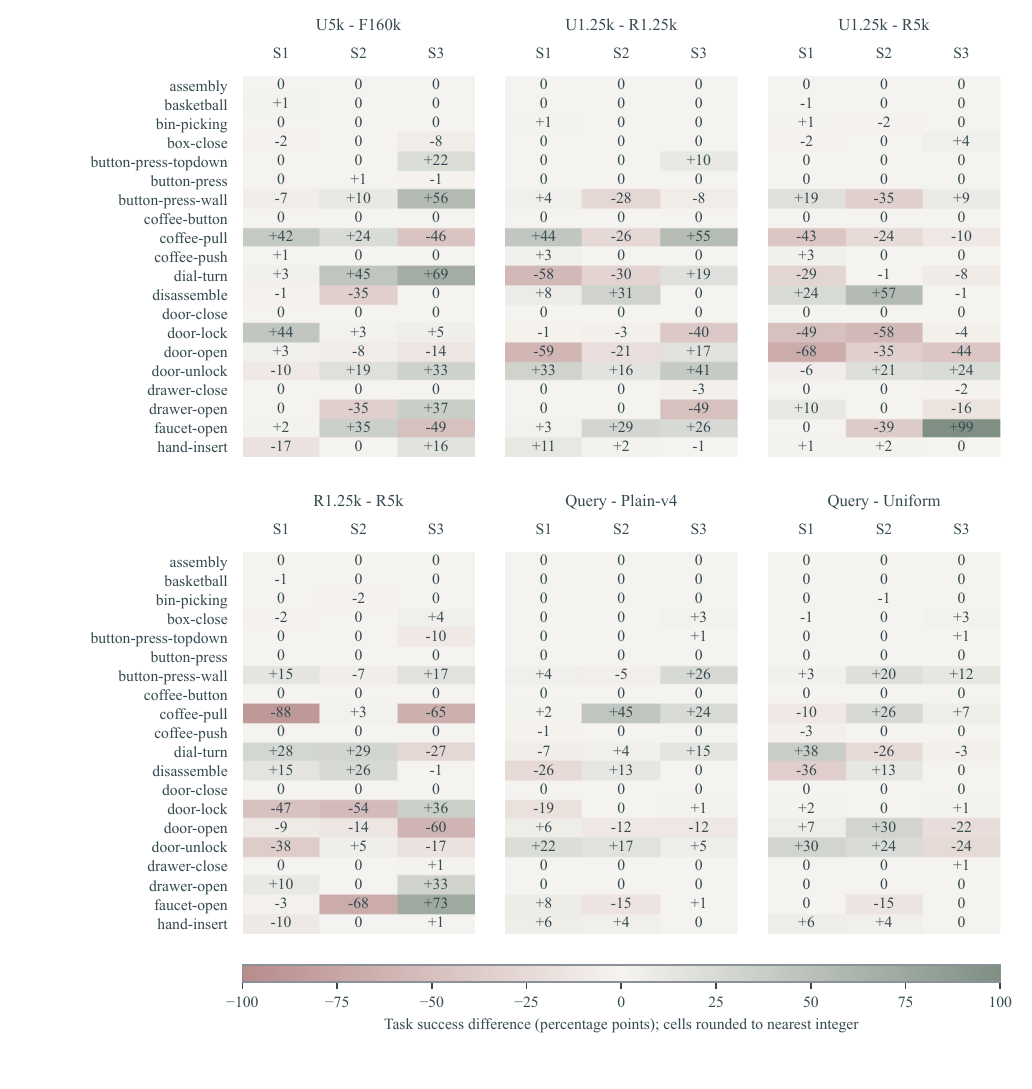}
\caption{Task-by-seed effects for the remaining six comparisons. The data-selection, compression, and routing families retain their own controls.}
\label{fig:heat2}
\end{figure*}
\clearpage

\onecolumn
\section{Development history and its role in the paper}
\label{app:history}
The project also explored alignment variants, combinations of selection scores, downstream update rules, auxiliary objectives, and small correction pilots. Table~\ref{tab:history} lists the recorded comparisons in chronological order with their original controls. Single-seed selection results that later became part of a three-seed study count only within that study.

The initial HTB5000 combination was $4.13\pp$ below Random5000, and equal weighting of the U/T/B/E criteria was $3.68\pp$ below its U-only reference. Several update variants were screened on seed 1. The late4 follow-up used robot training seeds 2 and 3 with the human parent held at seed 1; its mean difference was $-1.60\pp$, supplying two subsequent downstream repetitions from that single human parent.

Among the auxiliary objectives---chronology (D1), visual transfer (D2), causal prefix-to-future and physical targets (D3), task routing (D4), difference-based queries (D5), and synchrony (D6)---D1--D3, D5, and D6 scored below their designated controls, while D4 advanced to the three-seed and mismatch studies. These screens motivated the later questions; the main claims rest on the full matched comparisons.

The small correction pilots use a different intervention and evaluation unit. One 20-case paired correction test improved one case, left 17 unchanged, and worsened two. These case-level tests are reported separately from the 20-task macro comparisons, and historical rows that summarize the same contrast at different stages are labeled in the table and source data.

\begingroup
\scriptsize
\begin{longtable}{p{.22\linewidth}p{.34\linewidth}lrr}
\caption{Historical development contrasts with their designated controls. Training-seed identifiers distinguish individual late4 runs from the seed-2/3 summary; these rows are not independent experiments. D4 contrasts superseded by the three-seed confirmation are reported in the main comparison tables.}\label{tab:history}\\
\toprule
Variant & Designated control & Seeds & Blocks & $\Delta$ (pp) \\
\midrule
\endfirsthead
\toprule
Variant & Designated control & Seeds & Blocks & $\Delta$ (pp) \\
\midrule
\endhead
Q & H\_TBE-v4 & 1 & 1 & -1.77 \\
H\_TBE & H\_TBE-v4 & 1 & 1 & -1.42 \\
Q-v4-query & H\_TBE-v4 & 1 & 1 & -3.26 \\
Q-query-ensemble & H\_TBE-v4 & 1 & 1 & -9.76 \\
H-v4-shift50 & H\_TBE-v4 & 1 & 1 & +0.83 \\
H\_TBE-v1 & H\_TBE-v4 & 1 & 1 & -1.42 \\
U5000 & Full40000 & 1 & 1 & +6.53 \\
U5000 & Random5000 & 1 & 1 & +1.28 \\
HTB5000 & Full40000 & 1 & 1 & +1.11 \\
HTB5000 & Random5000 & 1 & 1 & -4.13 \\
HTB5000 & U5000 & 1 & 1 & -5.42 \\
equal-weight U+T+B+E & U-only v4 & 1 & 1 & -3.68 \\
u\_relative & matched U same training seed & 1 & 1 & -0.62 \\
consensus & matched U same training seed & 1 & 1 & -0.69 \\
consensus\_projected & matched U same training seed & 1 & 1 & -0.90 \\
block8 & matched U same training seed & 1 & 1 & -1.15 \\
late4 & matched U same training seed & 1 & 1 & -0.28 \\
late4 & matched U same training seed & 2 & 1 & -2.05 \\
late4 & matched U same training seed & 3 & 1 & -1.15 \\
late4 & matched U same training seed & 2;3 & 2 & -1.60 \\
task-gated U continuation & U-continuation same budget & 1 & 1 & -0.94 \\
D1-chronology & original H\_TBE-v4 U (seed1) & 1 & 1 & -3.37 \\
D2-visual-transfer & original H\_TBE-v4 U (seed1) & 1 & 1 & -10.52 \\
D3-C causal prefix-to-future & D3-C0 future-only same targets & 1 & 1 & -5.76 \\
D3-P physical wrist/aperture & original H\_TBE-v4 U seed1 & 1 & 1 & -5.28 \\
D5-Difference & D5-Full & 1 & 1 & -2.74 \\
D6-Sync & D6-Shuffled & 1 & 1 & -5.42 \\
\bottomrule
\end{longtable}

\endgroup

\clearpage
\section{Reproducibility and artifact scope}
\label{app:reproduction}
\paragraph{Result-level reproduction.}
The local analysis package contains compact saved outcome tables, the recorded comparison specification, cross-seed diagnostic outputs, and scripts for reconstructing the paper's numerical summaries and figures. Integer per-task success counts permit reconstruction of the three-block macro means, paired differences, and descriptive intervals. It also includes 13,824 saved simulator episode outcomes with matched reset identifiers. These reproduce the original 5,000-draw observation bootstrap: all 36 interval endpoints match the saved report exactly in the recorded analysis environment. Appendix~\ref{app:protocol} specifies the resampling algorithm. The original fixed cross-seed procedure is included in a path-portable form and checked against the saved analysis. These calculations require no training checkpoints, simulator, or GPU.

The data are partitioned by their source experimental families. Reused Random5000 endpoints are identified through their arm, training seed, and task, and are counted once when constructing prediction matrices. The observation and slot tables remain separate. Paper-table generation reads directly from the saved values and reconstructed statistics. Every figure has a source mapping and a local rendering command.

\paragraph{Training provenance and availability.}
The protocol appendix documents data windows, selectors, model interfaces, optimization, and evaluation, with compact parameters, normalization conventions, the slot mapping, and reference code excerpts in \texttt{protocol/}. Full implementation snapshots and saved configurations remain with the authors, and end-to-end retraining additionally requires the Mecka export, subset manifests, normalization arrays, robot demonstrations or their generation environment, model initialization, and compatible software dependencies.

Appendix~\ref{app:protocol} describes the November runtime restoration and input-stream verification. Newer archives are recorded in later delivery receipts; their remote availability has not been rechecked since delivery.

\paragraph{Data access and release scope.}
The local release candidate contains numerical simulator outcomes, study descriptions, and per-component third-party notices; it excludes human videos, trajectories, recording identifiers, normalization arrays, and model weights. \texttt{DATA-PROVENANCE.md} records the provenance evidence for the historical Mecka snapshot. Its acquisition terms and redistribution permissions, which current EgoVerse software licensing and public episode labels do not settle, and the release terms for study-owned code and results await confirmation before any public repository or archival deposit.

\paragraph{Experiment execution versus analysis.}
All figures and statistical summaries are computed from the completed logical endpoints, while the source experiment logs record run-level execution and recovery history.

\paragraph{Reproduction entry points.}
The package README gives dependency installation and paper-build commands. \texttt{reproduction/README.md} documents the saved-result analysis; \texttt{reproduction/FIGURE-SOURCES.md} maps figures and tables to their inputs. \texttt{ASSET-AVAILABILITY.md} distinguishes included results, reference code, reconstructed inputs, and missing or access-controlled assets. The data dictionary defines statistical units and the episode-outcome fields.

\end{document}